\documentclass[a4paper,11pt]{article}
\usepackage[T1]{fontenc} 
\usepackage{float} 

\makeatletter
\gdef\@fpheader{ }
\gdef\@journal{ }
\makeatother
\RequirePackage{amsmath}
\RequirePackage{amssymb}
\RequirePackage{epsfig}
\RequirePackage{graphicx}
\RequirePackage[numbers,sort&compress]{natbib}
\RequirePackage{color}
\RequirePackage[colorlinks=true
,urlcolor=blue
,anchorcolor=blue
,citecolor=blue
,filecolor=blue
,linkcolor=blue
,menucolor=blue
,pagecolor=blue
,linktocpage=true
,pdfproducer=medialab
,pdfa=true
]{hyperref}

\newif\ifnotoc\notocfalse
\newif\ifemailadd\emailaddfalse
\newif\iftoccontinuous\toccontinuousfalse
\makeatletter
\def\@subheader{\@empty}
\def\@keywords{\@empty}
\def\@abstract{\@empty}
\def\@xtum{\@empty}
\def\@dedicated{\@empty}
\def\@arxivnumber{\@empty}
\def\@collaboration{\@empty}
\def\@collaborationImg{\@empty}
\def\@proceeding{\@empty}
\def\@preprint{\@empty}

\newcommand{\subheader}[1]{\gdef\@subheader{#1}}
\newcommand{\keywords}[1]{\if!\@keywords!\gdef\@keywords{#1}\else%
\PackageWarningNoLine{\jname}{Keywords already defined.\MessageBreak Ignoring last definition.}\fi}
\renewcommand{\abstract}[1]{\gdef\@abstract{#1}}
\newcommand{\dedicated}[1]{\gdef\@dedicated{#1}}
\newcommand{\arxivnumber}[1]{\gdef\@arxivnumber{#1}}
\newcommand{\proceeding}[1]{\gdef\@proceeding{#1}}
\newcommand{\xtumfont}[1]{\textsc{#1}}
\newcommand{\correctionref}[3]{\gdef\@xtum{\xtumfont{#1} \href{#2}{#3}}}
\newcommand\jname{JHEP}

\newcommand\preprint[1]{\gdef\@preprint{\hfill #1}}

\makeatother

\newcommand\note[2][]{%
\if!#1!%
\stepcounter{footnote}\footnotetext{#2}%
\else%
{\renewcommand\thefootnote{#1}%
\footnotetext{#2}}%
\fi}

\makeatletter
\newtoks\auth@toks
\renewcommand{\author}[2][]{%
  \if!#1!%
    \auth@toks=\expandafter{\the\auth@toks#2\ }%
  \else
    \auth@toks=\expandafter{\the\auth@toks#2$^{#1}$\ }%
  \fi
}
\makeatother
\makeatletter
\newtoks\affil@toks\newif\ifaffil\affilfalse
\newcommand{\affiliation}[2][]{%
\affiltrue
  \if!#1!%
    \affil@toks=\expandafter{\the\affil@toks{\item[]#2}}%
  \else
    \affil@toks=\expandafter{\the\affil@toks{\item[$^{#1}$]#2}}%
  \fi
}
\makeatother
\makeatletter
\newtoks\email@toks\newcounter{email@counter}%
\newcommand{\emailAdd}[1]{%
\emailaddtrue%
\ifnum\theemail@counter>0\email@toks=\expandafter{\the\email@toks, \@email{#1}}%
\else\email@toks=\expandafter{\the\email@toks\@email{#1}}%
\fi\stepcounter{email@counter}}
\newcommand{\@email}[1]{\href{mailto:#1}{\tt #1}}
\makeatother

\makeatletter
\newcommand*\collaboration[1]{\gdef\@collaboration{#1}}
\newcommand*\collaborationImg[2][]{\gdef\@collaborationImg{#2}}
\makeatletter
\newcommand\afterLogoSpace{\smallskip}
\newcommand\afterSubheaderSpace{\vskip3pt plus 2pt minus 1pt}
\newcommand\afterProceedingsSpace{\vskip21pt plus0.4fil minus15pt}
\newcommand\afterTitleSpace{\vskip23pt plus0.06fil minus13pt}
\newcommand\afterRuleSpace{\vskip23pt plus0.06fil minus13pt}
\newcommand\afterCollaborationSpace{\vskip3pt plus 2pt minus 1pt}
\newcommand\afterCollaborationImgSpace{\vskip3pt plus 2pt minus 1pt}
\newcommand\afterAuthorSpace{\vskip5pt plus4pt minus4pt}
\newcommand\afterAffiliationSpace{\vskip3pt plus3pt}
\newcommand\afterEmailSpace{\vskip16pt plus9pt minus10pt\filbreak}
\newcommand\afterXtumSpace{\par\bigskip}
\newcommand\afterAbstractSpace{\vskip16pt plus9pt minus13pt}
\newcommand\afterKeywordsSpace{\vskip16pt plus9pt minus13pt}
\newcommand\afterArxivSpace{\vskip3pt plus0.01fil minus10pt}
\newcommand\afterDedicatedSpace{\vskip0pt plus0.01fil}
\newcommand\afterTocSpace{\bigskip\medskip}
\newcommand\afterTocRuleSpace{\bigskip\bigskip}
\newlength{\affiliationsSep}
\newcommand\beforetochook{\pagestyle{myplain}\pagenumbering{roman}}

\DeclareFixedFont\trfont{OT1}{phv}{b}{sc}{11}

\renewcommand\maketitle{
\pagestyle{empty}
\thispagestyle{titlepage}
\setcounter{page}{0}
\noindent{\small\scshape\@fpheader}\@preprint\par

\afterLogoSpace
\if!\@subheader!\else\noindent{\trfont{\@subheader}}\fi
\afterSubheaderSpace
\if!\@proceeding!\else\noindent{\sc\@proceeding}\fi
\afterProceedingsSpace
{\LARGE\flushleft\sffamily\bfseries\@title\par}
\afterTitleSpace
\hrule height 1.5\p@%
\afterRuleSpace
\if!\@collaboration!\else
{\Large\bfseries\sffamily\raggedright\@collaboration}\par
\afterCollaborationSpace
\fi
\if!\@collaborationImg!\else
{\normalsize\bfseries\sffamily\raggedright\@collaborationImg}\par
\afterCollaborationImgSpace
\fi
{\bfseries\raggedright\sffamily\the\auth@toks\par}
\afterAuthorSpace
\ifaffil\begin{list}{}{%
\setlength{\leftmargin}{0.28cm}%
\setlength{\labelsep}{0pt}%
\setlength{\itemsep}{\affiliationsSep}%
\setlength{\topsep}{-\parskip}}
\itshape\small%
\the\affil@toks
\end{list}\fi
\afterAffiliationSpace
\ifemailadd 
\noindent\hspace{0.28cm}\begin{minipage}[l]{.9\textwidth}
\begin{flushleft}
\textit{E-mail:} \the\email@toks
\end{flushleft}
\end{minipage}
\else 
\PackageWarningNoLine{\jname}{E-mails are missing.\MessageBreak Plese use \protect\emailAdd\space macro to provide e-mails.}
\fi
\afterEmailSpace
\if!\@xtum!\else\noindent{\@xtum}\afterXtumSpace\fi
\if!\@abstract!\else\noindent{\renewcommand\baselinestretch{.9}\textsc{Abstract:}}\ \@abstract\afterAbstractSpace\fi
\if!\@keywords!\else\noindent{\textsc{Keywords:}} \@keywords\afterKeywordsSpace\fi
\if!\@arxivnumber!\else\noindent{\textsc{ArXiv ePrint:}} \href{http://arxiv.org/abs/\@arxivnumber}{\@arxivnumber}\afterArxivSpace\fi
\if!\@dedicated!\else\vbox{\small\it\raggedleft\@dedicated}\afterDedicatedSpace\fi
\ifnotoc\else
\iftoccontinuous\else\newpage\fi
\beforetochook\hrule
\tableofcontents
\afterTocSpace
\hrule
\afterTocRuleSpace
\fi
\setcounter{footnote}{0}
\pagestyle{myplain}\pagenumbering{arabic}
} 

\renewcommand{\baselinestretch}{1.1}\normalsize
\divide\@tempdima\baselineskip
\@tempcnta=\@tempdima
\skip\@mpfootins = \skip\footins
\renewcommand{\@dotsep}{10000}

\newcommand\ps@myplain{
\pagenumbering{arabic}
\renewcommand\@oddfoot{\hfill-- \thepage\ --\hfill}
\renewcommand\@oddhead{}}
\let\ps@plain=\ps@myplain

\newcommand\ps@titlepage{\renewcommand\@oddfoot{}\renewcommand\@oddhead{}}

\numberwithin{equation}{section}

\renewcommand\section{\@startsection{section}{1}{\z@}%
                                   {-3.5ex \@plus -1.3ex \@minus -.7ex}%
                                   {2.3ex \@plus.4ex \@minus .4ex}%
                                   {\normalfont\large\bfseries}}
\renewcommand\subsection{\@startsection{subsection}{2}{\z@}%
                                   {-2.3ex\@plus -1ex \@minus -.5ex}%
                                   {1.2ex \@plus .3ex \@minus .3ex}%
                                   {\normalfont\normalsize\bfseries}}
\renewcommand\subsubsection{\@startsection{subsubsection}{3}{\z@}%
                                   {-2.3ex\@plus -1ex \@minus -.5ex}%
                                   {1ex \@plus .2ex \@minus .2ex}%
                                   {\normalfont\normalsize\bfseries}}
\renewcommand\paragraph{\@startsection{paragraph}{4}{\z@}%
                                   {1.75ex \@plus1ex \@minus.2ex}%
                                   {-1em}%
                                   {\normalfont\normalsize\bfseries}}
\renewcommand\subparagraph{\@startsection{subparagraph}{5}{\parindent}%
                                   {1.75ex \@plus1ex \@minus .2ex}%
                                   {-1em}%
                                   {\normalfont\normalsize\bfseries}}

\def\fnum@figure{\textbf{\figurename\nobreakspace\thefigure}}
\def\fnum@table{\textbf{\tablename\nobreakspace\thetable}}

\long\def\@makecaption#1#2{%
  \vskip\abovecaptionskip
  \sbox\@tempboxa{\small #1. #2}%
  \ifdim \wd\@tempboxa >\hsize
    \small #1. #2\par
  \else
    \global \@minipagefalse
    \hb@xt@\hsize{\hfil\box\@tempboxa\hfil}%
  \fi
  \vskip\belowcaptionskip}

\renewenvironment{thebibliography}[1]{%
\begin{oldthebibliography}{#1}%
\small%
\raggedright%
\setlength{\itemsep}{5pt plus 0.2ex minus 0.05ex}%
}%
{%
\end{oldthebibliography}%
}

\begin{document}


\title{\boldmath Activation functions are not needed: the ratio net}

\author[a,1]{Chi-Chun Zhou,}\note{zhouchichun@dali.edu.cn.  Corresponding author}
\author[b]{Hai-Long Tu,}\note{Hai-Long Tu and Chi-Chun Zhou contributed equally to this work.}
\author[a]{Yue-Jie Hou,}
\author[a]{Zhen Ling,}
\author[a]{Yi Liu,}\note{liuyi@dali.edu.cn. }
\author[a]{and Jian Hu}\note{hujian@dali.edu.cn.}

\affiliation[a]{School of Engineering, Dali University, Dali, Yunnan 671003, PR China}
\affiliation[b]{RoyalFlush Information Network Co.,Ltd, HangZhou, ZheJiang 310023, PR China}










\abstract{A deep neural network for classification
tasks is essentially consist of two components:  
feature extractors and function approximators.
They usually work as an integrated whole, 
however, improvements on any components
can promote the performance of the whole algorithm.
This paper focus on designing a new function approximator.
Conventionally, to build a function approximator, 
one usually uses the method based on the nonlinear
activation function or the nonlinear kernel function and yields classical networks
such as the feed-forward neural network (MLP) 
and the radial basis function network (RBF). 
In this paper, a new function approximator that is 
effective and efficient is proposed. 
Instead of designing new activation functions or kernel functions, 
the new proposed network uses the fractional form. 
For the sake of convenience, we name the network the ratio net. 
We compare the
effectiveness and efficiency of the ratio net and that of 
the RBF and the MLP with various kinds of activation functions  
in the classification task on the mnist database of 
handwritten digits and the Internet Movie Database (IMDb) which is a binary sentiment analysis dataset. 
It shows that, in most cases,
the ratio net converges faster and outperforms 
both the MLP and the RBF.}


\maketitle
\flushbottom


\section{Introduction}
A deep neural network for classification tasks 
is essentially consist of two components:
(1) feature extractors that extract the feature out from the raw data.
For example, the convolutional neural network (CNN)
\cite{o2015introduction,graves2013speech,mikolov2010recurrent} can effectively extract
usefull features from raw images,
the recurrent neural network (RNN) \cite{kim2014convolutional,hu2014convolutional}
can effectively extract hidden features from time series,
and the transformer \cite{vaswani2017attention} can effectively extract
context-features from raw texts.
And (2) function approximators that find the function mapping features to
labels. For example, the feed-forward neural network (MLP) 
\cite{white1990connectionist,hornik1990universal,leshno1993multilayer} 
is widely used as a function approximators in finding the target function mapping the 
multi-dimensional input-features to the multi-dimensional output-labels.
In a deep neural network, 
these two components usually work as an integrated whole, 
however,
any improvements on feature extractors or function approximators
can promote the performance of the whole algorithm.

In this paper, we focus on designing 
a new function approximator which has 
smaller number of parameters 
and faster training algorithms 
and at the same time performances better.

Conventionally, there are two ways to build a 
function approximator, (1) the nonlinear
activation function \cite{leshno1993multilayer,glorot2011deep,xu2015empirical,sonoda2017neural,misra2019mish}
and (2) the nonlinear kernel function \cite{park1991universal,chen1991orthogonal}.
Along these two roads, two types of classical neural networks, 
the MLP \cite{white1990connectionist,hornik1990universal,leshno1993multilayer} and 
the RBF \cite{park1991universal,chen1991orthogonal}, are proposed, respectively.
Moreover, to ensure the effectiveness, researchers proveded that
function approximators
should have the property of universal
approximation \cite{hornik1989multilayer,white1990connectionist,hornik1990universal,leshno1993multilayer,park1991universal}. 
That is, the target function should be approximated with an
arbitrarily small error with the function approximator.
And the known classical MLP and RBF are proved to have the property of universal
approximation \cite{hornik1989multilayer,white1990connectionist,hornik1990universal,
leshno1993multilayer,park1991universal,chen1991orthogonal}.

Along the road of activation function, attempts are made to find 
a better function approximator by introducing new activation functions,
For example, the swish activation function is provided by the Google Brain \cite{ramachandran2017searching}, 
the mish activation function is suggested in Ref. \cite{misra2019mish}, 
a fast exponentially linear unit (FELU)
is proposed in Ref. \cite{qiumei2019improved}, 
an improved rectified linear unit (relu)
segmentation correction activate function is proposed in Ref. \cite{lin2018research}, 
and a dynamic modification of 
the activation function is proposed in Ref. \cite{mercioni2019dynamic}.
The binarized neural networks with activations constrained to $+1$ or $-1$ 
is proposed in Ref. \cite{courbariaux2016binarized}. 
Moreover, comparisions and discussions on various kinds of activation functions 
attract researchers' attentions. For example, 
a comparision of the relu function and the sigmoid function is considered and
a self-adaptive evolutionary algorithm to search for activation functions is given in Ref. \cite{nader2020searching}. 
A comparison of the expressive power of deep neural networks (DNNs) with the relu 
activation function to linear spline methods is given in Ref. \cite{eckle2019comparison}.
A survey on the existing activation functions and the trends in the use
of the activation functions for deep learning applications is given in 
Refs. \cite{nwankpa2018activation,sharma2017activation}.
The standard relu, the leaky relu, the parametric relu, 
and the new randomized leaky relu activation functions 
are evaluated in image classification tasks \cite{xu2015empirical}.
A technique to automatic discover new activation functions is given in Ref. \cite{ramachandran2017searching}.
However, the strategy of weight initializations
and the choice of activation functions
have a non neglected impact on the performance of the training procedure \cite{hayou2019impact,ramachandran2017searching}. 

Along the road of nonlinear kernel function, online RBF algorithm which is coupled with a kernel principal component analysis
(KPCA) is proposed in Ref. \cite{errachdi2017online}.
Besides, there are discussions on designing a method that essentially 
acts as a function approximator finding relations between the multi-input and the multi-output 
in a nonlinear sysytem.
For example, adaptive switch controller (ASC) is designed for the onlinear multi-input
multi-output system (MIMO) in Ref. \cite{slama2019adaptive} and 
indirect adaptive control methods using neural network are proposed for nonlinear
systems in Refs. \cite{slama2019neural,errachdi2018performance}.

In this paper, 
we propose a new function approximator 
that is efficient in finding the
target function mapping the feature to the label. 
Unlike the conventional method, 
instead of designing a new nonlinear kernel function in the RBF or searching for 
new activation functions in the MLP, 
the new proposed network uses the fractional form. 
For the sake of convenience, we name the
network the ratio net. 
The ratio net is inspired by the previous work \cite{zhou2020pade}, where
we find that the Pad{\'e} approximant with fractional forms 
is highly efficient in searching a target
function. We compare the effectiveness and efficiency of the ratio net with 
classical networks, the MLP and the RBF, in the classification task on the mnist database of 
handwritten digits \cite{lecun1998mnist,deng2012mnist} and the IMDb \cite{dooms2013movietweetings} which 
is a binary sentiment analysis dataset.
The result shows that the ratio net converges faster and, in most cases, outperforms 
both the MLP with various kind of activation function and the RBF, even with smaller number of parameters.

The work is organized as follows: in Sec. 2, we give the details of the
ratio net. In Sec. 3, we compare the effectiveness and efficiency of the ratio
net with that of the classical networks such as the MLP and the RBF in the classification task on
the mnist database of handwritten digits and the IMDb. 
Conclusions and outlooks are given in Sec. 4.
Other details of the experiments are given in the appendix.
The source code of the present paper is given in 
\href{https://github.com/zhouchichun/implementation_of_the_ratio_net}{github} .

\section{The ratio net}

The nonlinearity between the multi-dimensional 
input-features and multi-dimensional output-labels
is the main difficulty in designing the function
approximator. 
In this section, instead of along the conventional roads,
we introduce a new way to handle the difficulty caused by the
nonlinearity: the ratio net with fractional forms. 
Here, we describe the structure, the training algorithm, 
and the universal approximation property of the ratio net.

\subsection{The structure} 

\begin{figure}[H]
\centering
\includegraphics[width=0.9\textwidth]{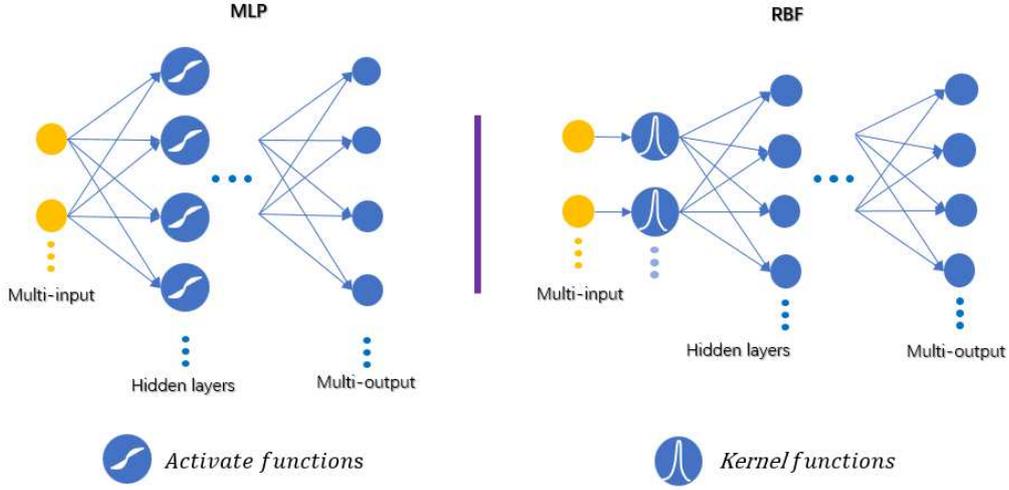}
\caption{An diagram showing the structure of the MLP and the RBF.}
\label{mlp}
\end{figure}

\textit{A brief review of the MLP and the RBF. }
The function approximator is essentially a function 
that maps a vector in the space $R^{N}$ to
another vector in the space $R^{M}$. 
Conventionally, to overcome the onlinearity between the multi-dimensional 
input-features and multi-dimensional output-labels, one usually uses the method based on the nonlinear
activation function or the nonlinear kernel function and yields classical networks
such as the MLP and the RBF. The structure of the single-layer MLP and the single-layer RBF reads:
 $\left.  y_{mlp}\left(  x\right)  \right\vert _{i}%
=w_{i}^{\prime}\sigma\left(  \sum_{j=1}^{n}w_{ij}x_{j}+b_{1}\right)  +b_{2}$
and $\left.  y_{rbf}\left(  x\right)  \right\vert _{i}=w_{i}^{\prime}k\left(
\left\vert x_{j}-c_{j}\right\vert \right)  +b_{2}$ where $\left.  y\left(
x\right)  \right\vert _{i}$ is the $ith$ component of the output,
$\sigma\left(  x\right)  $ is the activation function, and $k\left(  x\right)  $
is the kernel function. $w_{i}^{\prime}$, $w_{ij}$, $b_{l}$, and $c_{j}$ are
parameters. Usually, $\sigma\left(  x\right)  $ can be chosen from the sigmoid
function, the hyperbolic tangent (tanh) function, the relu function, and etc.. 
$k\left(  x\right)  $ can
be a Gauss function. As shown in Fig. (\ref{mlp}). 

\begin{figure}[H]
\centering
\includegraphics[width=0.9\textwidth]{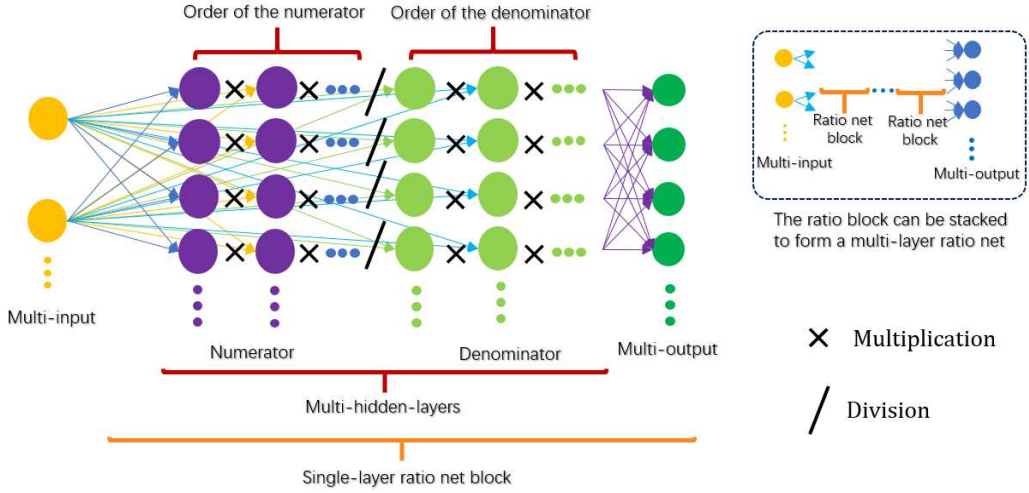}
\caption{An diagram showing the structure of the ratio net.}
\label{ratio}
\end{figure}

\textit{The structure of the ratio net. } 
Unlike the MLP and the RBF, the ratio net uses the fractional from 
to overcome the onlinearity.
The structure of the single-layer ratio net is
\begin{equation}
\left.  y_{rn}\left(  x\right)  \right\vert _{i}= \sum_{l=1}%
^{h}w_{il}\left[\frac{\left(  \sum_{j=1}%
^{n}w_{lj}^{\prime}x_{j}+b_{l}^{\prime}\right)  \left(  \sum_{j=1}^{n}%
w_{lj}^{\prime\prime}x_{j}+b_{l}^{\prime\prime}\right)  \ldots}{\left(
\sum_{j=1}^{n}w_{lj}^{\prime\prime\prime}x_{j}+b_{l}^{\prime\prime\prime
}\right)  \left(  \sum_{j=1}^{n}w_{lj}^{\prime\prime\prime\prime}x_{j}%
+b_{l}^{\prime\prime\prime\prime}\right)  \ldots}\right]+b_{i}, \label{Eq1}%
\end{equation}
where $w_{lj}$ and $b_{j}$ are parameters. 
$j$ runs from $1$ to $n$, the dimension of input,
$l$ runs from $1$ to $h$, the dimension of hidden layer,
and $i$ runs from $1$ to $m$, the dimension of output.
In Eq. (\ref{Eq1}), instead of the
nonlinear activation function or the nonlinear kernel function, 
the nonlinearity
is overcomed by using the fractional form. As shown in Fig. (\ref{ratio}). 

\begin{figure}[H] 
\centering
\includegraphics[width=0.9\textwidth]{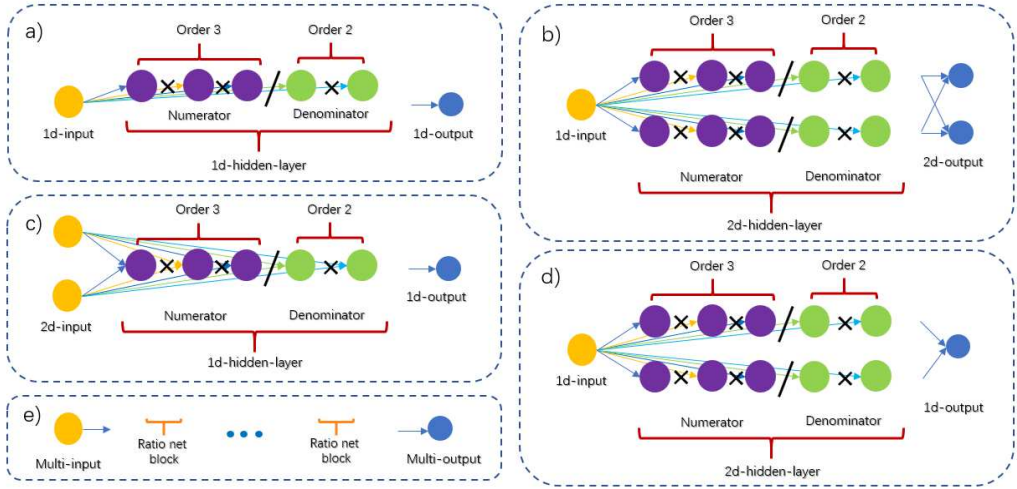}
\caption{An diagram showing the structure of the ratio net at simple cases.
$"a"$ shows the ratio net with $1$-$d$ input, $1$-$d$ output, and $1$-$d$ hidden layer,
$"b"$ shows the ratio net with $1$-$d$ input, $2$-$d$ output, and $2$-$d$ hidden layer, 
$"c"$ shows the ratio net with $2$-$d$ input, $1$-$d$ output, and $1$-$d$ hidden layer, and 
$"d"$ shows the ratio net with $1$-$d$ input, $1$-$d$ output, and $2$-$d$ hidden layer.
Those four cases are all single-layer ratio nets with
the order of numerator $3$ and the order of denominator $2$.
$"e"$ shows multi-layer ratio nets with multi-input, multi-output, and multi-hidden layer.}
\label{ratio_simple}
\end{figure}

To better illustrate the structure of the ratio net, Fig. (\ref{ratio_simple})
shows four simple cases. They are $1$-$d$ input to $1$-$d$ output with $1$-$d$ hidden layer, 
$1$-$d$ input to $2$-$d$ output with $2$-$d$ hidden layer, 
$2$-$d$ input to $1$-$d$ output with $1$-$d$ hidden layer, and
$1$-$d$ input to $1$-$d$ output with $2$-$d$ hidden layer.
Those four cases are all ratio nets with single hidden layer, 
the order of numerator $3$, and the order of denominator $2$.

We specify the case with $1$-$d$ input, $1$-$d$ output, and $2$-$d$ hidden 
layer by the following equation,
\begin{equation}
y_{rn}\left(  x\right)  =\sum_{l=1}^{2}w_{l}\left[  \frac{\left(
w_{l}^{\prime}x+b^{\prime}\right)  \left(  w_{l}^{\prime\prime}x+b^{\prime
\prime}\right)  \left(  w_{l}^{\prime\prime\prime}x+b^{\prime\prime\prime
}\right)  }{\left(  w_{l}^{\prime\prime\prime\prime}x+b^{\prime\prime
\prime\prime}\right)  \left(  w_{l}^{\prime\prime\prime\prime\prime
}x+b^{\prime\prime\prime\prime\prime}\right)  }\right]  +b. \label{Eq11}%
\end{equation}
Fig. (\ref{ratio_simple}) shows the structure of these simple cases of the ratio net.

To conclude, a ratio net is decided by the following hyper parameters: 
the order of denominator,
the order of numerator, 
and the hidden-layer size provided the dimension 
of the input and the output is known.
Although, the single-layer ratio net can be stacked to form a multi-layer ratio net,
the single-layer ratio net already performs well.
Like most neural networks, the optimal hyper parameters
can be decided under trials.

\subsection{The training algorithm}
The parameters, $w_{ij}$ and $b_{j}$, in the ratio net, Eq. (\ref{Eq1}),
are updated by the back propagation algorithm \cite{lecun1988theoretical,cilimkovic2015neural}.
The updating of the parameters
in a ratio net can be automatically implemented by open source deep learning frameworks,
such as the Tensorflow \cite{abadi2016tensorflow} used in the present work.

\subsection{The property of universal approximation of the ratio
net: a brief discussion} 
In designing the function approximator, 
one has to make sure that the target
function mapping the feature to the label is in the range of searching. 
That is, 
the network should have the property of universal approximation.
Although, a rigorous proof of the property of universal
approximation of the ratio net is not given, we show that the property of universal approximation
of the ratio net is inherited from the Pad{\'e} approximant \cite{baker1964theory,baker1996pade}.

The conventional Pad{\'e} approximant is given as \cite{baker1964theory,baker1996pade}
\begin{equation}
y_{pade}\left(  x\right)  =\frac{\sum_{l=0}^{L}a_{l}x^{l}}{\sum_{j=0}^{M}%
b_{j}x^{j}}%
\end{equation}
with $L$ and $M$ integers.
The Pad{\'e} approximant has a Maclaurin expansion which agrees with
a power series of order $L+M+1$ \cite{baker1996pade}, that is
\begin{equation}
y_{pade}\left(  x\right)  \sim\sum_{l=0}^{L+M+1}c_{l}x^{l}.%
\end{equation}
Thus, the Pad{\'e} approximant 
is capable of approximating various kind of functions. 

The conventional Pad{\'e} approximant gives functions that map a one-dimension
input to another one-dimension output. 
The ratio net proposed in the paper is
a generalization of the Pad{\'e} approximant and gives functions that map a vector in
$R^{N}$ to another vector in $R^{M}$. 
The property of universal approximation
of the ratio net is inherited from the Pad{\'e} approximant. 

\section{The classification task on the mnist database of handwritten digits and the IMDb}

In this section, we compare the
effectiveness and efficiency of the ratio net and that of the classical networks such
as the MLP and the RBF on the classification task on the mnist database of handwritten digits and the IMDb. 
It shows that the ratio net with smaller number 
of parameters converges faster and, in most cases, 
outperforms the MLP and the RBF .

\subsection{The task on the mnist database of handwritten digits}

\begin{figure}[H]
\centering
\includegraphics[width=0.9\textwidth]{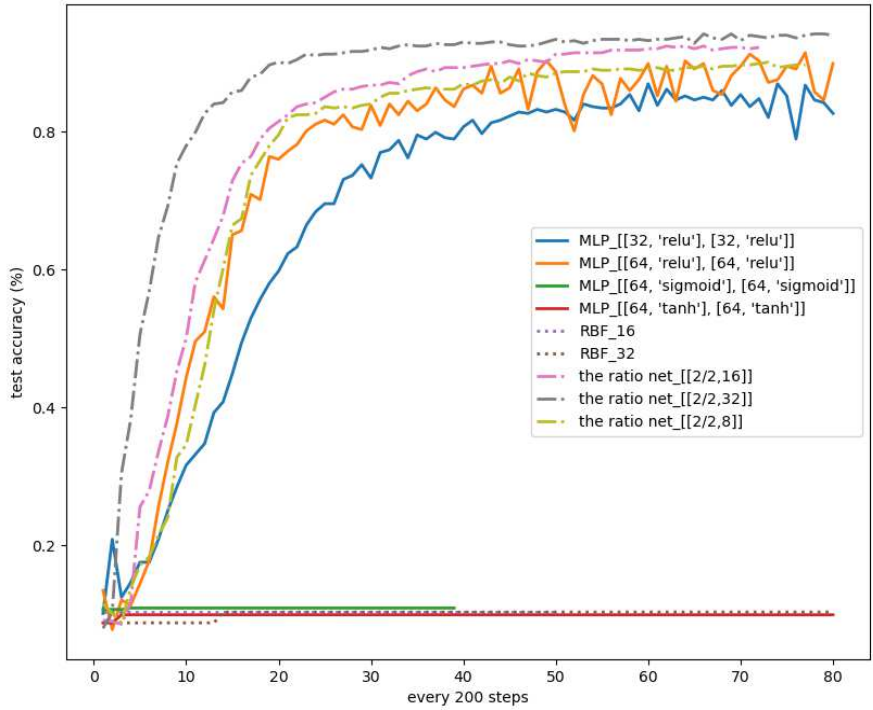}
\caption{The accuracy of different methods on the test set of the mnist database of handwritten digits versus the steps. 
The learning rates are all $0.0001$.}
\label{ex5}
\end{figure}

The mnist database of handwritten digits \cite{lecun1998mnist,deng2012mnist} is an famous open dataset for
image classification tasks. There are
$60,000$ training images and $10,000$ test
images in the database. 
In this section, we firstly use a
convolutional auto-encoder network (CAE) \cite{masci2011stacked} to extract features from a raw picture. 
The detail of the CAE is given in the appendix. As a
result, we obtained a $20$-dimensional feature from a raw picture with size $28\times28$. 
Then, the MLP, the RBF, and the ratio net with different structures are used to find the
function that maps the $20$-dimensional feature to the label, which has $10$
different values. 
The cross entropy \cite{mannor2005cross} between the predicted label and the real label is considered as the loss function.
The Adam algorithm \cite{kingma2014adam} is used to minimize the loss function.
The learning rates of different networks are all $0.0001$.
In the training proccess, the early stopping strategy \cite{caruana2001overfitting} is applied. 
That is, the training proccess stops if no
improvements are given on the accuracy of training set.
Results of different representative networks are given in Table. (\ref{table43}) and Fig. (\ref{ex5}).
\begin{table}[H]
\centering
\caption{The accuracy of different representative networks 
on the test set of the mnist database of handwritten digits.
The early stopping strategy is applied.}
\label{table43}
 \begin{tabular}{lll}  
\hline   
$\text{Structures }$& $\text{Number of parameters }$  &$\text{The accuracy}$\\  
\hline   
$\text{the ratio net: }[[2/2,8]] $& $762$ & $89.64$\\
$\text{the ratio net: }[[2/2,16]] $& $1,514$ & $92.18$\\
$\text{the ratio net: }[[2/2,32]] $& $3,018$ & $93.94$\\
$\text{MLP: }[[32,relu],[32,relu]]$&$ 2,058$ & $84.76$\\
$\text{MLP: }[[64,relu],[64,relu]]$&$ 6,154$ & $84.96$\\
$\text{MLP: }[[64,sigmoid],[64,sigmoid]]$&$ 6,154$ & $10.93$(fail)\\
$\text{MLP: }[[64,tanh],[64,tanh]]$&$ 6,154$ & $0.09$(fail)\\
$\text{RBF: }16$&$ 506$ & $0.09$(fail)\\
$\text{RBF: }32$&$ 1002$ & $0.09$(fail)\\
\hline 
\end{tabular}
\end{table}

In Table. (\ref{table43}), "the ratio net: $[[2/2,8]]$" denotes a single-layer ratio net with the order of numerator $2$, 
the order of denominator $2$, and 
the hidden-layer size $8$. 
We specify the corresponding structure with the equation, 
\begin{equation}
\left.  y_{rn}\left(  x\right)  \right\vert _{i}= \sum_{l=1}^{8}w_{il}\left[\frac{\left(  \sum_{j=1}%
^{20}w_{lj}^{\prime}x_{j}+b_{1}^{\prime}\right)  \left(  \sum_{j=1}^{20}%
w_{lj}^{\prime\prime}x_{j}+b_{1}^{\prime\prime}\right)}{\left(
\sum_{j=1}^{20}w_{lj}^{\prime\prime\prime}x_{j}+b_{1}^{\prime\prime\prime
}\right)  \left(  \sum_{j=1}^{20}w_{lj}^{\prime\prime\prime\prime}x_{j}%
+b_{1}^{\prime\prime\prime\prime}\right)}\right]+b_{i}, \label{Eq2}%
\end{equation}
where $x_{j}$ with $j$ runs from $1$ to $20$, the dimension of the features
and $i$ runs from $1$ to $10$, the dimension of the labels. 
For the sake of clarity, the dimensions of input and output are omitted.
"MLP:$[[64,tanh],[64,tanh]]$" denotes a two-layered MLP with $64$
neurons in each layer and the activation function the tanh function.
"RBF:$16$" denotes a single-layer RBF with hidden-layer size $16$.

The efficiency of each networks is shown in Fig. (\ref{ex5})
where the growing tendency of the accuracy on the test set versus the steps is given. 
It shows that the ratio net with smaller number of parameters
converges faster and outperforms the MLP and the RBF. 

\begin{figure}[H]
\centering
\includegraphics[width=0.9\textwidth]{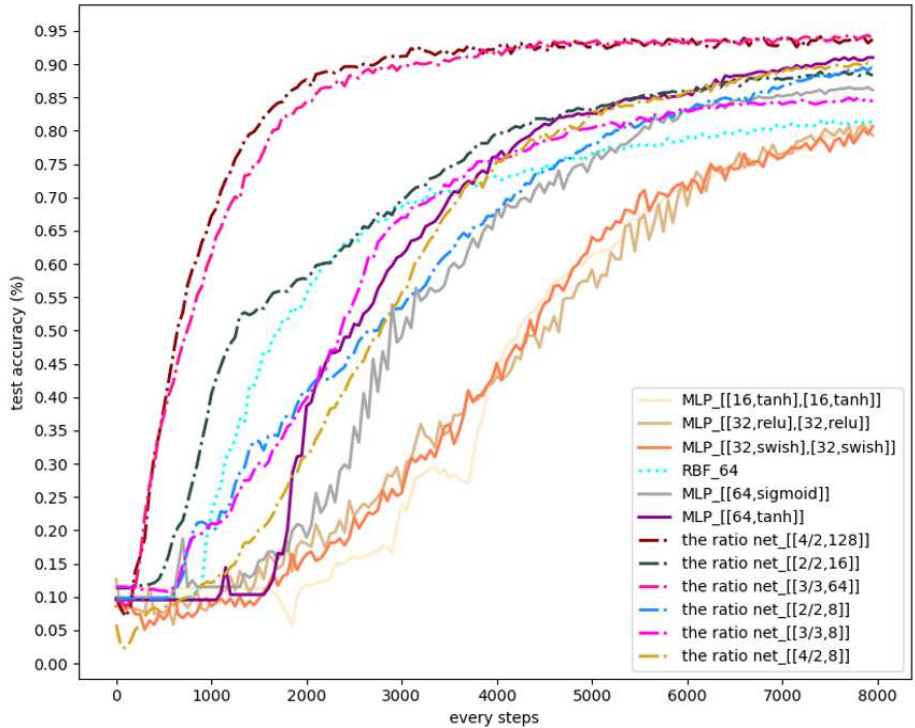}
\caption{The accuracy of different methods on the test set of 
the mnist database of handwritten digits versus the steps. 
The $20$-dimensional features are preproceed by the maximum and minimum normalization.
The learning rates are all $0.0001$.}
\label{ex5_new}
\end{figure}

Here is an extra experiment on the mnist database of handwritten digits
where the $20$-dimensional features are preproceed by the maximum and minimum normalization.
The result is shown in Table. (\ref{table4_new}) and Fig. (\ref{ex5_new}).
It shows that after maximum and minimum normalization preprocess,
the performances of the ratio net, the MLP, and the RBF are all improved.
Especially for the single-layer MLP with tanh functions and sigmoid functions.
The single-layer MLP with tanh function at this case outperforms the ratio net,
however, for multi-layer MLP with tanh functions and sigmoid functions, they 
sometimes fail in the task, as shown in Table. (\ref{table4_new}).

\begin{table}[H]
\centering
\caption{The accuracy of different representative networks 
on the test set of the mnist database of handwritten digits.
The $20$-dimensional features are preproceed by the maximum and minimum normalization.
The early stopping strategy is applied.}
\label{table4_new}
\begin{tabular}{lll}  
\hline   
 $\text{Structures }$& $\text{Number of parameters }$  &$\text{The accuracy}$\\  

\hline   
$\text{the ratio net: }[[2/2,8]] $& $762$ & $91.79$\\
$\text{the ratio net: }[[4/2,8]] $& $1,098$ & $91.21$\\
$\text{the ratio net: }[[3/3,16]] $& $2,186$ & $92.96$\\
$\text{the ratio net: }[[3/3,32]] $& $4,362$ & $94.14$\\
$\text{the ratio net: }[[3/3,64]] $& $8,714$ & $95.11$\\
$\text{the ratio net: }[[4/2,128]] $& $17,418$ & $95.50$\\
$\text{MLP: }[[64,tanh]]$&$ 1,994$ & $93.55$\\
$\text{MLP: }[[64,swish]]$&$ 1,994$ & $89.64$\\
$\text{MLP: }[[64,sigmoid]]$&$ 1,994$ & $90.82$\\
$\text{MLP: }[[32,relu],[32,relu]]$&$ 2,058$ & $89.84$\\
$\text{MLP: }[[64,tanh],[64,tanh]]$&$ 6,154$ & $11.32$(fail)\\
$\text{MLP: }[[64,tanh],[64,tanh],[64,tanh]]$&$ 10,314$ & $10.54$(fail)\\
$\text{MLP: }[[64,sigmoid],[64,sigmoid]]$&$6,154$ & $11.91$(fail)\\
$\text{MLP: }[[64,swish],[64,swish]]$&$ 6,154$ & $91.60$\\
$\text{RBF: }16$&$ 506$ & $69.33$\\
$\text{RBF: }32$&$ 1,002$ & $85.15$\\
$\text{RBF: }64$&$ 1,994$ & $87.30$\\

\hline 
\end{tabular}
\end{table}

It shows that the ratio net again
converges faster and, in most cases, outperforms the MLP and the RBF.

\subsection{The task on the IMDb}

\begin{figure}[H]
\centering
\includegraphics[width=0.9\textwidth]{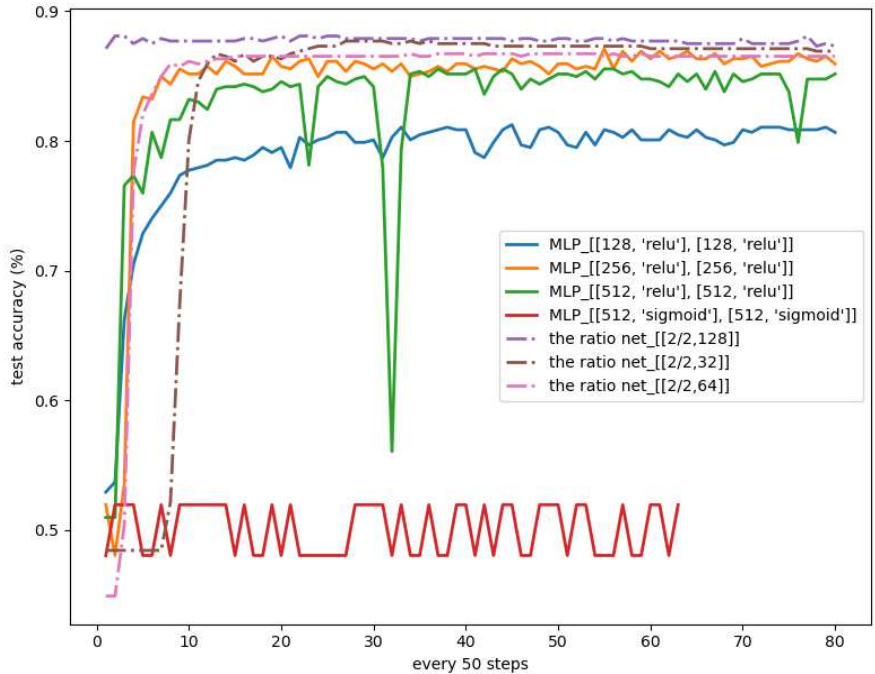}
\caption{The accuracy of different methods on the test set of the 
IMDb versus the steps. 
The learning rates are all $0.0001$.}
\label{ex5_imbd}
\end{figure}

The IMDb \cite{weible2001internet} is a binary sentiment analysis dataset consisting of $50,000$ reviews 
labeled as positive or negative.
The IMDb is an famous open dataset for nature language processing (NLP) classification tasks \cite{weible2001internet}. 
In this section, we use the
textcnn model \cite{kim2014convolutional} to extract features from a sentence. 
The detail of the textcnn model is given in the appendix.
As a result, we obtained a
$384$-dimensional feature from a given text sample in the IMDb. Then, the ratio net, 
the MLP, and the RBF with different structures are used to find the function that
maps the $384$-dimensional feature to the binary label. 
The cross entropy \cite{mannor2005cross} between the predicted label and the real label is considered as the loss function.
The Adam algorithm \cite{kingma2014adam} is used to minimize the loss function.
The learning rates of different networks are all $0.0001$.
In the training proccess, the early stopping strategy \cite{caruana2001overfitting} is applied. 
In the experiment, 
$1,000$ reviews are randomly selected from the standard test set of the IMDb and used as the new test set.
The result is given in Table. (\ref{tableimdb}) and Fig. (\ref{ex5_imbd}).
\begin{table}[H]
\centering
\caption{The accuracy of different representative networks 
on the test set of the IMDb.
The early stopping strategy is applied.}
\label{tableimdb} 
\begin{tabular}{lll} 
\hline   
  $\text{Structures }$& $\text{Number of parameters }$  &$\text{The accuracy}$\\  

\hline   
$\text{the ratio net: }[2/2,32] $& $49,346$ & $86.91$\\
$\text{the ratio net: }[2/2,64] $& $98,690$ & $86.52$\\
$\text{the ratio net: }[2/2,128] $& $197,378$ & $87.69$\\
$\text{MLP: }[[128,relu],[128,relu]]$&$ 66,050$ & $81.05$\\
$\text{MLP: }[[256,relu],[256,relu]]$&$ 164,866$ & $86.05$\\
$\text{MLP: }[[512,relu],[512,relu]]$&$ 460,802$ & $84.96$\\
$\text{text-cnn with random embedding}$&$-$ & $84.43$\\
\hline 
\end{tabular}
\end{table}

It also shows that on the task of the IMDb, the ratio net
converges faster and outperforms the MLP and the RBF.

\section{Conclusions and outlooks}

In this paper, we propose a new network, a function approximator, 
that is efficient in finding the
function that maps features to labels. Instead of using the nonlinear
activation functions or kernel functions, 
the new proposed network uses the fractional form to
overcome the nonlinearity between the multi-input and the multi-output. 
We compare the effectiveness and efficiency of the ratio net and that of the
classical networks such as the MLP and the RBF in the classification task 
on the mnist database of handwritten digits and the IMDb.
The result shows that the ratio net converges faster and, in most cases, 
outperforms both the MLP and the RBF even with smaller number of parameters.
The ratio net can be further used in supervised or unsupervised machine learning
tasks.

\section{Acknowledgments}
We are very indebted to Prof. Wu-Sheng Dai for his enlightenment and encouragement. 
We are very indebted to Prof. Guan-Wen Fang for his encouragement. 
This work is supported by National Natural Science Funds of China (Grant No. 62106033), 
Yunnan Youth Basic Research Projects (202001AU070020), 
and Doctoral Programs of Dali University (KYBS201910).

\section{Conflict of interest statement}
We declare that we have no financial 
and personal relationships with other people 
or organizations that can inappropriately influence our work, 
there is no professional or other personal interest of any 
nature or kind in any product, service and/or company that 
could be construed as influencing the position presented in, 
or the review of, the manuscript entitled, 
“Activation functions are not needed: the ratio net”.

\section{Appendix}
\subsection{The CAE used to encode the mnist database}
The CAE is an effective unsupervised machine learning method 
that can extract useful features from raw images \cite{masci2011stacked}.
The structure of the CAE used to extract features from the raw images 
of handwritten digits is given in Table. (\ref{CAE}).
The target layer gives the $20$-dimensional features.

\subsection{The textcnn used to encode the IMDb}
The textcnn is a useful feature extractor that can 
extract important features out of a sentence \cite{kim2014convolutional}.
The structure of the textcnn used to extract the features from the raw sentences 
of the IMDb database
is given in Table. (\ref{different_model_textcnn}).
The target layer gives the $384$-dimensional features.

\begin{table}[H]
\centering
\caption{The structure of the CAE.}
\label{CAE}
\begin{tabular}{|c|c|c|c|}
\hline
Layer  & Operation, kernel size, chanel &Input& Output \\ \hline
1       & input &28$\times$28& 28$\times$28$\times$1  \\ \hline
2      & Convolution, [3,3], 128&28$\times$28$\times$1 &28$\times$28$\times$128 \\ \hline
3      & Maxpooling, [2,2], -&28$\times$28$\times$128&14$\times$14$\times$128 \\ \hline
4      & Convolution, [3,3], 128&14$\times$14$\times$128&14$\times$14$\times$128 \\ \hline
5      & Maxpooling, [2,2], -&14$\times$14$\times$128&7$\times$7$\times$128 \\ \hline
6(target layer)       & Reshape and dense, [6272,20],-&7$\times$7$\times$128&20  \\ \hline
7       & Dense and reshape, [20,6272]&20&7$\times$7$\times$128   \\ \hline
8  &Up sampling, [2,2], -&7$\times$7$\times$128&14$\times$14$\times$128 \\ \hline
9       & Deconvolution, [3,3], 128&7$\times$7$\times$128&14$\times$14$\times$128  \\ \hline
10  &Up sampling, [2,2], -&14$\times$14$\times$128&28$\times$28$\times$128 \\ \hline
11       & Deconvolution, [3,3], 1&28$\times$28$\times$128&28$\times$28$\times$1  \\ \hline

\end{tabular}

\end{table}

\begin{table}[H]
\centering
\caption{The structure of the textcnn.}
\label{different_model_textcnn}
\begin{tabular}{|c|c|c|c|}
\hline
Layer  & Operation, size, chanel &Input& Output \\ \hline
1       & input &512& 512 \\ \hline
2      & Embedding and expand, 128,- &512&512$\times$128$\times$1 \\ \hline
3-1      & Convolution, [3,128], 128 &512$\times$128$\times$1&510$\times$1$\times$128\\ \hline
3-2      & Convolution, [4,128], 128&512$\times$128$\times$1&509$\times$1$\times$128 \\ \hline
3-3      & Convolution, [5,128], 128&512$\times$128$\times$1&508$\times$1$\times$128 \\ \hline
4-1       & Maxpooling, [510,1],-&510$\times$1$\times$128&128  \\ \hline
4-2       & Maxpooling, [509,1],-&509$\times$1$\times$128&128  \\ \hline
4-3       & Maxpooling, [508,1],-&508$\times$1$\times$128&128  \\ \hline
5(target layer)       & Concatenation,-, -&-&384   \\ \hline
6  &Dense, [384,2], -&384 &2  \\ \hline

\end{tabular}

\end{table}
\vspace{-1em}











\end{document}